%% file: AggCLIP_OneThree.tex
\PassOptionsToPackage{table}{xcolor}
\PassOptionsToPackage{normalem}{ulem}
\documentclass[twocolumn,letterpaper]{onethree}
\usepackage{amsmath,amssymb,pifont}
\usepackage{xurl}
\newcommand{\Description}[1]{}
\hypersetup{
  pdftitle={AggCLIP: Aggregating CLIP for Zero-Shot Multimodal Anomaly Detection},
  pdfauthor={Zhi Li, Hanyu Wang, Fei Shen, Zhengtao Zhang},
  pdfsubject={Computing methodologies: Visual inspection},
  pdfkeywords={Zero-Shot Anomaly Detection, Multimodal Anomaly Detection, CLIP, Prompt Learning}
}

\title{AggCLIP: Aggregating CLIP for Zero-Shot\\Multimodal Anomaly Detection}
\author[1,2]{\href{https://orcid.org/0009-0008-8633-690X}{Zhi Li}}
\author[1,2]{\href{https://orcid.org/0009-0007-7710-6523}{Hanyu Wang}}
\author[1,2,3,*]{\href{https://orcid.org/0000-0001-9263-4489}{Fei Shen}}
\author[1,2,3]{\href{https://orcid.org/0000-0003-1659-7879}{Zhengtao Zhang}}
\affiliation[1]{Institute of Automation, Chinese Academy of Sciences, Beijing, China}
\affiliation[2]{School of Artificial Intelligence, University of Chinese Academy of Sciences, Beijing, China}
\affiliation[3]{CASI Vision Technology Co., Ltd., Luoyang, Henan, China}
\contribution[*]{Corresponding author: \email{fei.shen@ia.ac.cn}}
\abstract{Zero-Shot Multimodal Anomaly Detection (ZSMAD) aims to identify both appearance and geometric defects in novel categories without relying on target-specific training data, making it especially valuable in real-world scenarios constrained by data scarcity, privacy concerns, or high annotation costs. While recent approaches successfully adapt vision-language models like CLIP to achieve this open-vocabulary generalization, they still face significant challenges: (i) severe geometric information loss during 3D-to-2D point cloud rendering; (ii) spatially noisy 2D feature maps produced by raw CLIP patch embeddings, compounded by shallow late-stage modality fusion; and (iii) severe cross-modal semantic misalignment due to decoupled processing streams. To address these limitations, we propose \textbf{AggCLIP}, a novel framework that adapts frozen CLIP representations for ZSMAD by explicitly incorporating local geometric structure and visual spatial context. Specifically, to accurately localize fine-grained defects, we design a Geometry-Aware Spatial Aggregation (GSA) module that refines back-projected 3D point features via density-adaptive neighborhood weighting, alongside a Context-Aware Spatial Aggregation (CSA) module that smooths multi-scale 2D patch embeddings using a lightweight trainable adapter. Furthermore, to prevent semantic drift between modalities, we introduce a Dual-Stream Prompt Contrastive (DSPC) loss, which aligns the abstract concepts of normality and abnormality across the point cloud and RGB streams. Extensive experiments demonstrate that the proposed AggCLIP framework achieves state-of-the-art performance on the MVTec 3D-AD and Eyecandies benchmarks under both one-vs-rest and cross-dataset zero-shot settings, notably surpassing the strongest competitive methods by up to 2.9\% in pixel-level AUPRO.}
\checkdata[Conference]{Author version of the ACM Multimedia 2026 paper.
  \href{https://doi.org/10.1145/3767308.3835551}{doi:10.1145/3767308.3835551}}
\checkdata[License]{\textcopyright{} 2026 the authors.
  \href{https://creativecommons.org/licenses/by/4.0/}{CC BY 4.0}. Layout adapted for arXiv.}

\begin{document}
\maketitle
\begin{figure*}[t]
    \centering
    \includegraphics[width=\linewidth]{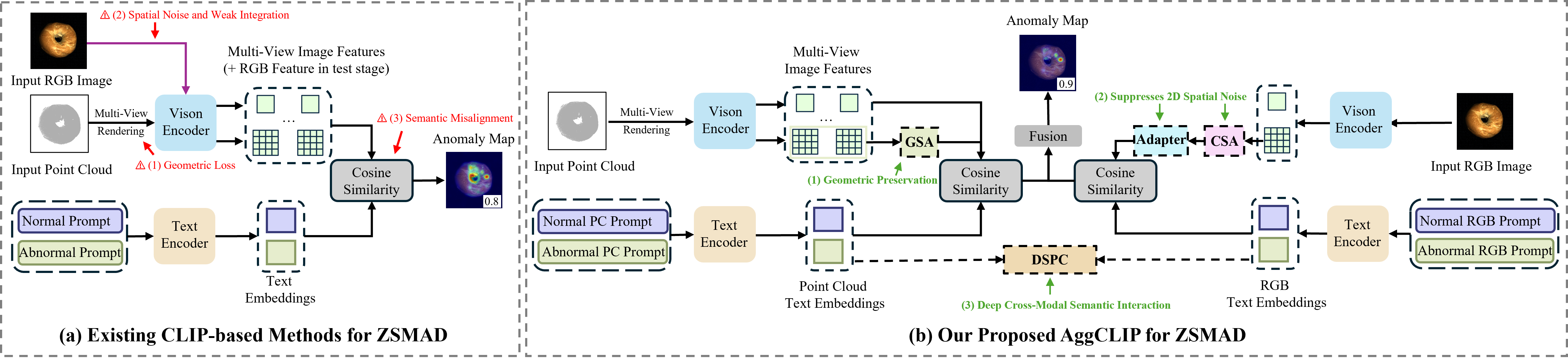}
    \caption{Paradigm comparison for zero-shot multimodal anomaly detection. \textbf{(a) Existing ZSMAD pipelines} typically suffer from three fundamental limitations: \textit{Geometric Loss} during 3D-to-2D rendering, \textit{Spatial Noise and Weak Integration} due to raw CLIP embeddings and simplistic late-fusion, and \textit{Semantic Misalignment} between prompts and multimodal features. \textbf{(b) The proposed AggCLIP framework} effectively addresses these gaps. It refines noisy back-projected point features via the Geometry-Aware Spatial Aggregation (\textbf{GSA}) module, suppresses 2D spatial noise using Context-Aware Spatial Aggregation (\textbf{CSA}) paired with a learnable \textbf{Adapter}, and directly enforces deep cross-modal semantic interaction via the Dual-Stream Prompt Contrastive (\textbf{DSPC}) loss.}
    \Description{A side-by-side schematic contrasts existing zero-shot multimodal anomaly detection pipelines with AggCLIP. The existing pipeline highlights geometric information loss, spatial noise, weak modality fusion, and semantic misalignment. AggCLIP adds GSA for point features, CSA and an adapter for RGB features, and DSPC for cross-modal prompt alignment.}
    \label{fig:methods_compare}
\end{figure*}

\section{Introduction}
\label{sec:intro}

Industrial anomaly detection (IAD) is a critical process in modern manufacturing, playing a key role in ensuring stable operations and minimizing economic losses~\cite{mvtec-ad,patchcore,rd4ad}. Previous methods have primarily focused on RGB images for detection~\cite{padim,patchcore,draem,dinomaly}. While these methods are effective for discovering texturing and surface defects, they inherently fail to capture structural and shape anomalies that are indiscernible in 2D projections. Consequently, modern industrial inspection systems increasingly demand the integration of paired RGB images and 3D point clouds for Multimodal Anomaly Detection (MAD).

Traditional MAD methods~\cite{m3dm,cfm,mmrd,shape-guided,pirn,bridgenet,cmdiad,unimmad,ast} typically rely on unsupervised learning schemas, which require training on a massive amount of defect-free samples from the target category. However, in real-world scenarios, acquiring sufficient training data for novel or diverse target categories is often difficult due to issues like commercial confidentiality, data privacy concerns, and prohibitive annotation costs. To address these challenges, Zero-Shot Multimodal Anomaly Detection (ZSMAD) has emerged as an essential task, aiming to identify and localize geometric and appearance defects on unseen product categories without relying on any target-specific training data.

CLIP~\cite{CLIP}, pre-trained on hundreds of millions of image-text pairs, establishes a powerful joint vision-language representation space with strong generalizability. Prompted by this capability, recent studies~\cite{PointAD,GS-CLIP,ZUMA} have applied CLIP to zero-shot multimodal anomaly detection. The predominant paradigm involves rendering 3D point clouds into multi-view 2D images, processing these images with CLIP to derive anomaly concepts, and subsequently back-projecting the outputs to 3D space. However, as illustrated in Figure~\ref{fig:methods_compare}, existing ZSMAD methods still face three major limitations:

\textbf{1) Geometric Information Loss.} Projecting point clouds into 2D views discards fine-grained 3D structure, forcing CLIP to represent structural anomalies through 2D visual proxies and limiting its sensitivity to subtle geometric defects.

\textbf{2) Spatial Noise and Superficial Modality Integration.} Existing methods often discard paired RGB images or fuse them only at the final stage. Moreover, raw CLIP patch embeddings lack local spatial inductive biases, producing fragmented and noisy maps for dense anomaly localization.

\textbf{3) Cross-Modal Semantic Misalignment.} Processing multi-modal visual inputs without an explicit semantic regularization mechanism inevitably causes domain drift. This leads to a severe misalignment in the conceptual understanding of ``normal'' and ``abnormal'' boundaries between the text prompt space and the decoupled RGB and point cloud representation streams.

To overcome these limitations, we propose \textbf{AggCLIP}, a simple yet highly effective framework for zero-shot multimodal anomaly detection. Our key insight is that effective zero-shot inference requires explicitly incorporating local geometric structure and visual spatial context into frozen CLIP representations while strictly preserving cross-modal semantic consistency. Towards this end, we design two modality-specific spatial aggregation modules tailored for anomaly perception. For the 3D branch, we introduce the \textbf{Geometry-Aware Spatial Aggregation (GSA)} module, which refines noisy back-projected point features using density-adaptive neighborhoods defined on the original 3D coordinates. For the 2D branch, we design the \textbf{Context-Aware Spatial Aggregation (CSA)} module, paired with a lightweight adapter, to smooth hierarchical 2D patch embeddings and suppress spatial noise. Furthermore, to mitigate semantic drift, we propose a \textbf{Dual-Stream Prompt Contrastive (DSPC)} loss, which dynamically regularizes the textual prompt learners to rigorously align the abstractions of normality and abnormality across both modalities.

Our main contributions are summarized as follows:
\begin{itemize}
    \item We propose \textbf{AggCLIP}, a universally applicable zero-shot multimodal anomaly detection framework that effectively bridges frozen CLIP representations with unified 2D appearance and 3D geometric reasoning.
    \item We design GSA and CSA to refine back-projected point features with local 3D neighborhood context and RGB patch features with multi-scale spatial context, producing spatially coherent anomaly predictions.
    \item We introduce the Dual-Stream Prompt Contrastive (DSPC) loss, which explicitly enforces semantic alignment between the decoupled text prompt spaces and the multimodal visual feature streams.
    \item Extensive experiments on the MVTec 3D-AD and Eyecandies demonstrate that AggCLIP establishes new state-of-the-art object- and pixel-level performance under both one-vs-rest and challenging cross-dataset zero-shot settings.
\end{itemize}

\section{Related Work}
\label{sec:related_work}

\subsection{Classical Anomaly Detection}

\textbf{2D Anomaly Detection.} Image-based industrial anomaly detection has been studied extensively under unsupervised and one-class settings. Existing methods can be broadly categorized into three major paradigms based on their underlying mechanisms. \textit{Feature embedding-based} methods~\cite{padim,patchcore,cflow,DFM} construct compact representations of normal patterns extracted from pre-trained networks, subsequently detecting anomalies by measuring structural deviations against memory banks or learned generative density models. \textit{Reconstruction-based} methods~\cite{rd4ad,stfpm,efficientad,uniad,dinomaly,dinomaly2,mambaad,inp-former,omnial,decad} leverage autoencoders and knowledge-distillation architectures trained exclusively on normal data, identifying anomalous regions by exploiting high spatial reconstruction errors. Furthermore, \textit{synthesis-based} methods~\cite{draem,destseg,RealNet,GLASS} artificially generate pseudo-anomalous training samples through procedural texture pasting or diffusion-based synthesis, thereby enabling discriminative classifiers to learn explicit decision boundaries. Despite their widespread success, these approaches operate entirely on 2D RGB images and are inherently incapable of capturing geometric structural defects that remain invisible in planar textures.

\noindent\textbf{Multimodal Anomaly Detection.} To mitigate the limitations of single-modality perception, the release of comprehensive large-scale datasets such as MVTec 3D-AD~\cite{mvtec-3dad} and Eyecandies~\cite{eyecandies} has driven rapid progress in multimodal anomaly detection. By jointly integrating 2D texture features and 3D spatial information, this paradigm enables a highly enriched and effective representation of complex physical anomaly patterns. Representative approaches have explored a diverse array of cross-modal fusion strategies. These include augmenting classical handcrafted 3D descriptors with learned 2D features~\cite{btf}, conducting bidirectional cross-modal feature mapping~\cite{cfm,m3dm}, applying reconstruction-style multimodal distillation~\cite{mmrd}, and leveraging prototype and shape-aware 3D geometric reasoning~\cite{shape-guided,pirn,bridgenet,ast,cmdiad,unimmad}. While demonstrating strong in-domain detection capabilities, these conventional multimodal frameworks remain fundamentally dependent on a massive corpus of defect-free normal samples from the target category for unsupervised training. In real-world industrial environments, strict constraints involving data privacy boundaries and prohibitive collection costs often make it exceptionally difficult to acquire sufficient normal samples. This intrinsic bottleneck significantly restricts their scalability, directly motivating the urgent exploration of zero-shot paradigms.

\subsection{Zero-Shot Anomaly Detection}

\textbf{Zero-Shot 2D Anomaly Detection.} ZSAD on unseen categories has become feasible by leveraging the generalized image-text alignment capabilities of CLIP~\cite{CLIP,CLIP-AD,SimCLIP,MuSc,InCTRL,AA-CLIP}. WinCLIP~\cite{WinCLIP} first applied CLIP to ZSAD tasks by aggregating sliding window features with compositional text prompts. Subsequently, various prompt learning frameworks~\cite{AnomalyCLIP,AdaCLIP,VCP-CLIP,BayesPFL} replaced static templates with learnable, object-agnostic, or context-aware tokens to adaptively capture localized defect semantics. However, these methods operate exclusively on planar 2D inputs, inherently restricting their capability to detect stereoscopic structural anomalies.

\noindent\textbf{Zero-Shot Multimodal Anomaly Detection.} Extending ZSAD to multimodal inputs aims to jointly localize both geometric and appearance defects without any target-specific training data. PointCLIP~\cite{PointCLIP,PointCLIPv2} established the standard paradigm of rendering 3D point clouds into multi-view images for CLIP processing and back-projecting the features to 3D space. Building on this, PointAD~\cite{PointAD} first adapted this paradigm to zero-shot anomaly detection by evaluating 3D structures from both point and pixel perspectives. Subsequent models like GS-CLIP~\cite{GS-CLIP} and ZUMA~\cite{ZUMA} further improved this pipeline via geometry-aware prompt regularization and training-free multimodal scoring, respectively. A fundamental limitation shared by these frameworks is their superficial incorporation of the native RGB modality and the total absence of explicit cross-modal semantic alignment. Our AggCLIP addresses this by incorporating local geometric structure and spatial context via modality-specific spatial aggregation modules (GSA and CSA) and enforcing multimodal semantic consistency via the DSPC loss.

\begin{figure*}[t]
    \centering
    \includegraphics[width=0.95\linewidth]{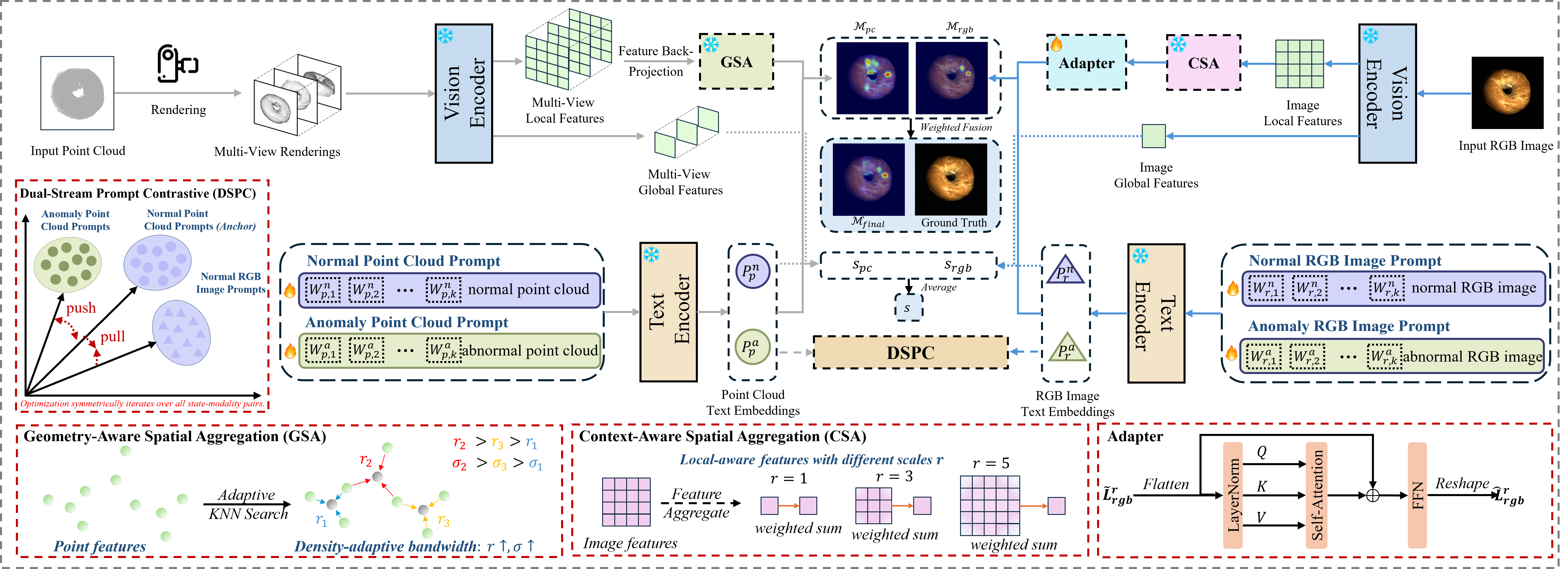}
    \caption{\textbf{Overview of AggCLIP.} GSA refines back-projected point-cloud features, CSA and a lightweight adapter refine RGB patch features, and DSPC aligns modality-specific text prompt embeddings before multimodal anomaly scoring.}
    \Description{The AggCLIP pipeline processes rendered point-cloud views and an RGB image with a shared frozen CLIP encoder. GSA aggregates the back-projected 3D features, while CSA and a trainable adapter refine the RGB features. Dual-stream prompt contrastive learning aligns normal and abnormal concepts before the two branches produce classification scores and anomaly maps.}
    \label{fig:overview}
\end{figure*}

\section{Problem Formulation}
\label{sec:problem_formulation}

Given a multimodal input $\mathcal{X}=(\mathcal{P}, \mathcal{I})$, we aim to learn a unified model $f_\theta: \mathcal{X} \to (s, \hat{\mathcal{M}})$ that jointly predicts a sample-level anomaly score $s \in [0, 1]$ and a pixel-level anomaly map $\hat{\mathcal{M}} \in [0, 1]^{H \times W}$. Here, the input consists of a 3D point cloud $\mathcal{P} \in \mathbb{R}^{M \times 3}$ and a geometrically registered RGB image $\mathcal{I} \in \mathbb{R}^{H \times W \times 3}$. Sample-level anomaly detection can be formally defined as a binary classification problem, where the model aims to classify samples $\mathcal{X}$ as either normal ($y=0$) or abnormal ($y=1$). Anomaly segmentation extends this concept to the pixel level, utilizing a binary ground truth mask $\mathcal{M} \in \{0, 1\}^{H \times W}$. We operate within the Zero-Shot Anomaly Detection (ZSAD) context. Crucially, ``zero-shot'' here refers to the \emph{target product categories}, meaning that the model has access to zero labeled examples from the target categories during training, but it is \emph{not} training-free. Specifically, we train on an auxiliary dataset $\mathcal{D}_{\text{train}} = \{(\mathcal{X}_i, \mathcal{M}_i, y_i)\}_{i=1}^N$ whose categories are entirely disjoint from those in the target testing set $\mathcal{D}_{\text{test}}$, i.e., $\mathcal{C}_{\text{train}} \cap \mathcal{C}_{\text{test}} = \emptyset$. Despite the distribution shift $P(\mathcal{X}_{\text{test}}) \neq P(\mathcal{X}_{\text{train}})$, the model is optimized on $\mathcal{D}_{\text{train}}$ to generalize to unseen categories in $\mathcal{D}_{\text{test}}$.

\section{AggCLIP}
\label{sec:methods}

\subsection{Overview}
\label{subsec:overview}

As illustrated in Figure~\ref{fig:overview}, we propose AggCLIP, a unified multimodal framework specifically designed for zero-shot 2D RGB and 3D point cloud anomaly detection. The primary objective of AggCLIP is to adapt the open-vocabulary generalized representations of a frozen vision-language model (CLIP) to the fine-grained demands of multimodal anomaly segmentation.

The AggCLIP framework operates in three sequential phases. First, we independently extract base appearance and geometric representations by passing the native RGB image and multi-view rendered point clouds through a shared, frozen CLIP vision encoder (Sec.~\ref{subsec:feature_extraction}). Second, to compensate for the inherent lack of local spatial inductive biases in raw CLIP features, we refine these embeddings via two modality-specific modules: the Geometry-Aware Spatial Aggregation (GSA) module for density-adaptive refinement of back-projected point features, and the Context-Aware Spatial Aggregation (CSA) module paired with a trainable adapter for 2D spatial noise suppression (Sec.~\ref{subsec:spatial_aggregation}). Finally, to prevent semantic drift between the isolated visual representations during adaptation, we introduce decoupled prompt learners regulated by a novel Dual-Stream Prompt Contrastive (DSPC) loss. This explicitly enforces cross-modal consistency by aligning the modality-specific text prompt embeddings for normality and abnormality across the point cloud and RGB streams (Sec.~\ref{subsec:prompt_learning}). By orchestrating these components around a frozen backbone, AggCLIP produces fine-grained anomaly score maps via text-visual similarity computation, enabling accurate zero-shot anomaly localization.

\subsection{Multimodal Feature Extraction}
\label{subsec:feature_extraction}

\subsubsection{Point Cloud Modality Feature Extraction}

To leverage the open-vocabulary priors of CLIP for 3D data without training a separate 3D encoder, we adopt the \textit{Render-Extract-Backproject} strategy. Given an input point cloud $\mathcal{P} \in \mathbb{R}^{M \times 3}$, we place $K$ virtual cameras around the object and render the point cloud from each viewpoint $k$ into an image $\mathcal{I}_k = \mathcal{R}(\mathcal{P}, k)$, converting 3D geometric structures into 2D visual representations. Each rendered image is fed into the frozen CLIP encoder $\Phi_{\text{img}}$ to extract local features $\mathbf{L}_k \in \mathbb{R}^{h' \times w' \times D}$ and global features $\mathbf{g}_k \in \mathbb{R}^D$. The per-point base features $\mathbf{F}_{\mathcal{P}}^{\text{base}} \in \mathbb{R}^{M \times D}$ are then obtained by back-projecting $\mathbf{L}_k$ onto each 3D point via the pinhole camera model, sampling with bilinear interpolation, filtering occluded views with a Z-buffer visibility test, and averaging the valid features across all viewpoints (full derivation in the Appendix).

\subsubsection{RGB Image Modality Feature Extraction}

The input RGB image $\mathcal{I}_{\text{rgb}} \in \mathbb{R}^{H \times W \times 3}$ is processed by the same frozen encoder $\Phi_{\text{img}}$ to obtain $\{\mathbf{g}_{\text{rgb}}, \mathbf{L}_{\text{rgb}}\} = \Phi_{\text{img}}(\mathcal{I}_{\text{rgb}})$, where $\mathbf{g}_{\text{rgb}} \in \mathbb{R}^{D}$ and $\mathbf{L}_{\text{rgb}} \in \mathbb{R}^{h' \times w' \times D}$ denote the global and local features, respectively. Sharing $\Phi_{\text{img}}$ across modalities ensures that RGB and rendered point cloud features reside in the same semantic space.

\subsection{Multimodal Feature Spatial Aggregation}
\label{subsec:spatial_aggregation}

A fundamental challenge in adapting CLIP for zero-shot anomaly detection is that its features, inherently trained for image-level alignment, are noisy and lack fine-grained spatial or geometric awareness. To bridge this gap, we design two core components of our \textbf{AggCLIP} framework: Geometry-Aware Spatial Aggregation (GSA) for the 3D modality and Context-Aware Spatial Aggregation (CSA) for the 2D modality. These modules explicitly incorporate local geometric structure and spatial context into the frozen CLIP features.

\subsubsection{Geometry-Aware Spatial Aggregation (GSA)}

The projected 3D point features $\mathbf{F}_{\mathcal{P}}^{\text{base}}$ initially represent isolated semantic observations, lacking geometric context. Anomalies on 3D surfaces are structurally defined by their deviations from local neighborhoods. Furthermore, raw point clouds exhibit highly non-uniform density, being dense at sharp edges and sparse on flat surfaces. Traditional fixed-scale aggregation struggles with this variance. To address this, GSA introduces a \textbf{Density-Adaptive Geometry Enhancement} mechanism.

First, we construct a $k$-Nearest Neighbor ($k$-NN) graph on the 3D coordinates $p_i \in \mathbb{R}^3$ and compute a local sparsity factor $\rho_i = \frac{1}{k} \sum_{j \in \mathcal{N}_k(i)} \|p_i - p_j\|_2$, i.e., the average Euclidean distance to its $k$ spatial neighbors. GSA then derives a density-adaptive aggregation bandwidth $\sigma_i = \gamma \cdot \rho_i$, where $\gamma$ is a scaling coefficient. This bandwidth shrinks the receptive field in dense regions to preserve fine defect details and expands it in sparse regions to gather sufficient structural context.

We then aggregate neighborhood features with a residual connection. The unnormalized aggregated feature $\tilde{\mathbf{f}}_i$ for point $p_i$ is:
\begin{equation}
    \tilde{\mathbf{f}}_i = \mathbf{f}_i^{\text{base}} + \sum_{j \in \mathcal{N}_k(i)} \frac{\exp\left( - \frac{\|p_i - p_j\|_2^2}{2\sigma_i^2} \right)}{\Omega_i} \cdot \mathbf{f}_j^{\text{base}}
\end{equation}
where $\Omega_i$ is the sum of the Gaussian kernel values over $\mathcal{N}_k(i)$, normalizing the weights into a distribution so that the aggregation term forms a convex combination of neighbor features. To ensure the aggregated features remain aligned with the pre-trained CLIP normalized embedding space, we apply $\ell_2$-normalization to the output: $\mathbf{f}_i^{\text{refined}} = \tilde{\mathbf{f}}_i / \| \tilde{\mathbf{f}}_i \|_2$. This yields the $\ell_2$-normalized refined features $\mathbf{F}_{\mathcal{P}}^{\text{refined}}$, which promote local 3D neighborhood consistency while retaining the original back-projected features through the residual connection.

\subsubsection{Context-Aware Spatial Aggregation (CSA)}

While the 2D local features $\mathbf{L}_{\text{rgb}}$ capture rich appearance semantics, the patch-based processing of the ViT backbone~\cite{vit} lacks the translation invariance and local inductive biases inherent in CNNs. In pixel-level tasks such as anomaly segmentation, this directly causes ``blocky'' artifacts and disjointed anomaly scores. To enforce spatial continuity without retraining the backbone, we propose the CSA module.

CSA employs a multi-scale smoothing strategy directly on the patch-resolution feature tokens across a predefined set of kernel sizes $\mathcal{S} = \{\kappa_1, \dots, \kappa_n\}$. For a given scale $\kappa$, the contextually aggregated feature at position $(u, v)$ on the patch grid is computed as:
\begin{equation}
    \begin{aligned}
    \widetilde{\mathbf{L}}_{\text{rgb}}^{\kappa}(u, v)
      &= \sum_{(\Delta u, \Delta v) \in \mathcal{R}_\kappa} \mathcal{G}(\Delta u, \Delta v; \sigma) \\
      &\quad {}\cdot \mathbf{L}_{\text{rgb}}(u+\Delta u, v+\Delta v)
    \end{aligned}
\end{equation}
where $\mathcal{G}(\cdot; \sigma)$ is a fixed Gaussian kernel shared across all scales and $\mathcal{R}_\kappa$ defines the local window of size $\kappa \times \kappa$. This parameter-free operation effectively mitigates local noise and ensures cohesive defect boundaries.

Although the aforementioned smoothing enforces spatial continuity, both the CLIP backbone and the Gaussian kernels remain completely frozen. To introduce trainable capacity without eroding the open-vocabulary behavior of the frozen backbone, we introduce the \textbf{Anomaly-Aware Patch Adapter}, a lightweight self-attention module that re-projects smoothed features into an anomaly-discriminative space. The aggregated features from all scales share one Adapter instance. We flatten $\widetilde{\mathbf{L}}_{\text{rgb}}^{\kappa}$ into a token sequence $\mathbf{Z}^{\kappa} \in \mathbb{R}^{N \times D}$ ($N = h' \times w'$) and project it into queries, keys, and values:
\begin{equation}
    \mathbf{Q}^\kappa = \mathbf{Z}^\kappa\mathbf{W}_Q, \quad \mathbf{K}^\kappa = \mathbf{Z}^\kappa\mathbf{W}_K, \quad \mathbf{V}^\kappa = \mathbf{Z}^\kappa\mathbf{W}_V
\end{equation}
where $\mathbf{W}_Q, \mathbf{W}_K, \mathbf{W}_V \in \mathbb{R}^{D \times D}$ are learnable projections following Layer Normalization. The adapted features are then obtained via self-attention with a residual connection and a subsequent feed-forward network:
\begin{equation}
    \begin{aligned}
    \hat{\mathbf{L}}_{\text{rgb}}^{\kappa}
      &= \text{Reshape}\bigl(\text{FFN}\bigl(\mathbf{Z}^\kappa \\
      &\qquad {}+ \text{Attention}(\mathbf{Q}^\kappa, \mathbf{K}^\kappa, \mathbf{V}^\kappa)\bigr)\bigr)
    \end{aligned}
\end{equation}
where $\text{FFN}(\cdot)$ is a two-layer feed-forward network with GELU activation. Here, the self-attention layer models global inter-patch dependencies, and the subsequent FFN maps the enriched features into a lower-dimensional anomaly-discriminative subspace.

Finally, we fuse the multi-scale adapted features in the \emph{similarity space}. For each scale $\kappa$, we compute the raw cosine similarity between the adapted patch features and the text embeddings $P_r^s$ (defined in Sec.~\ref{subsec:prompt_learning}), and accumulate these similarities across all scales before applying a single softmax:
\begin{equation}
    \begin{gathered}
    \mathbf{S}_{\text{rgb}}(u,v) = {}\\[-0.3ex]
    \text{softmax}_{s \in \{n, a\}}\left( \frac{1}{\tau_{vla}} \sum_{\kappa \in \mathcal{S}} \langle \hat{\mathbf{L}}_{\text{rgb}}^{\kappa}(u,v),\; P_r^s \rangle \right)
    \end{gathered}
\end{equation}
where $\tau_{vla}$ is a temperature parameter shared across all visual-language alignment computations. Applying the softmax across the semantic state $s \in \{n, a\}$ models the binary anomaly probability. This late-fusion strategy allows each scale to contribute its raw discriminative signal before normalization, preventing premature probability saturation that would occur if softmax were applied per-scale. Together, the Adapter serves as a lightweight domain-transfer bridge, where the self-attention layer captures inter-patch spatial dependencies and the FFN re-projects the contextualized features into an anomaly-aligned space for downstream multimodal fusion and scoring.

\subsection{Dual-Stream Prompt Contrastive Learning}
\label{subsec:prompt_learning}

In multimodal zero-shot anomaly detection, a critical challenge is maintaining consistent semantic boundaries for ``normal'' and ``abnormal'' across physically distinct modality representations, given that 3D geometric anomalies and 2D appearance anomalies manifest with vastly different visual patterns. Existing prompt-based methods typically adopt one of two strategies: either employing a single shared prompt template for all modalities, which cannot adequately capture the distinct characteristics of geometric versus appearance defects; or optimizing separate prompt learners for each modality in complete isolation without any cross-modal regularization. In the latter case, the independently learned semantic representations inevitably drift apart. Specifically, the 3D branch may associate ``anomaly'' predominantly with geometric deformations, while the 2D branch anchors it to textural irregularities. Both strategies ultimately lead to conflicting anomaly predictions during multimodal fusion, degrading detection and localization performance.

To address this, we introduce a Dual-Stream Prompt Contrastive (DSPC) learning strategy that decouples modality-specific prompts while encouraging cross-modal semantic consistency. We initialize independent sets of learnable context vectors for each modality. Let $m \in \{p, r\}$ denote the modality (where $p$ represents Point Cloud and $r$ represents RGB image), and $s \in \{n, a\}$ denote the semantic state (normal and abnormal, respectively). The prompt template for a specific modality and state is formulated as:
\begin{equation}
    T_m^s = [W_{m,1}^s][W_{m,2}^s]\dots [W_{m,L}^s] \oplus \text{state}^s \oplus \mathbf{c}_{m}
\end{equation}
where $W_{m,k}^s$ are the learnable context tokens, $L$ is the context length, $\text{state}^s \in \{\text{``normal''}, \text{``abnormal''}\}$ denotes the semantic state keyword, and $\mathbf{c}_{m} \in \{\text{``point cloud''}, \text{``RGB image''}\}$ is the modality identifier. By appending both the state label and the modality tag, the final prompt reads naturally as, e.g., ``\textit{[context] normal point cloud}'', grounding the text embedding to its corresponding visual branch. This dual-stream design allows each branch to learn modality-specific context representations while sharing a common state vocabulary.

\noindent \textbf{Dual-Stream Prompt Contrastive (DSPC) Loss.}
While decoupled learners provide domain flexibility, independent optimization may cause the cross-modal semantic representations to gradually drift apart. To regularize this, we propose the DSPC loss. Let $P_m^s$ denote the $\ell_2$-normalized text embedding generated by the frozen CLIP text encoder for a given prompt $T_m^s$. We construct positive pairs from cross-modal prompts sharing the same semantic state: $(P_p^n, P_r^n)$ and $(P_p^a, P_r^a)$. Negative pairs include both cross-modal conflicting states $(P_p^n, P_r^a)$, $(P_p^a, P_r^n)$ and intra-modality opposite states $(P_m^s, P_m^{\neg s})$.

To ensure fully symmetric alignment, we let each modality take turns as the anchor. Letting $\phi(a,b) = \langle a, b \rangle / \tau_{dspc}$ where $\langle \cdot , \cdot \rangle$ denotes the cosine similarity and $\tau_{dspc}$ is the temperature, we define the per-anchor contrastive term for modality $m$ with state $s$ as:
\begin{equation}
    \ell_m^s = -\log \frac{e^{\phi(P_m^s, P_{\bar{m}}^s)}}{e^{\phi(P_m^s, P_{\bar{m}}^s)} + e^{\phi(P_m^s, P_{\bar{m}}^{\neg s})} + e^{\phi(P_m^s, P_m^{\neg s})}}
\end{equation}
where $\bar{m}$ denotes the complementary modality. For each anchor $P_m^s$, the denominator contains three terms: (i) the positive $P_{\bar{m}}^{s}$ (same state, cross-modal), (ii) the cross-modal negative $P_{\bar{m}}^{\neg s}$ (opposite state, cross-modal), and (iii) the intra-modal negative $P_m^{\neg s}$ (opposite state, same modality). Together, terms (ii) and (iii) enforce cross-modal semantic consistency and sharpen the intra-modal normal--abnormal decision boundary. The full DSPC loss averages over all four anchor combinations:
\begin{equation}
    \mathcal{L}_{\text{dspc}} = \frac{1}{4} \sum_{m \in \{p, r\}} \sum_{s \in \{n, a\}} \ell_m^s
\end{equation}
This bidirectional formulation guarantees that both the point cloud and RGB branches are equally regularized: each modality's intra-modal normal--abnormal boundary is explicitly sharpened, and the cross-modal semantic alignment is enforced symmetrically.

\subsection{Training Objectives and Inference}

\noindent \textbf{Anomaly Score Map.}
For a target instance $(\mathcal{P}, \mathcal{I})$, the global sample-level classification score $s$ is obtained by averaging the softmax-normalized probabilities from the global visual features of both modalities:
\begin{equation}
    \begin{aligned}
    s &= \frac{1}{2}( s_{r} + s_{p} ), \\
    s_{m} &= \frac{1}{|\mathcal{G}_m|} \sum_{\mathbf{g} \in \mathcal{G}_m} \frac{\exp(\langle \mathbf{g}, P_m^a \rangle / \tau_{vla})}{\sum_{s \in \{n,a\}} \exp(\langle \mathbf{g}, P_m^s \rangle / \tau_{vla})}
    \end{aligned}
\end{equation}
where $\mathcal{G}_r = \{\mathbf{g}_{\text{rgb}}\}$ for the RGB image and $\mathcal{G}_p = \{\mathbf{g}_k\}_{k=1}^K$ for the $K$ rendered 3D views.

Simultaneously, the dense anomaly probability map $\mathbf{p}_m^a$ is derived analogously using the local aggregated visual features $v_{m,i}$, where $v_{m,i}$ denotes the feature vector at the $i$-th spatial position from the GSA-refined point features ($m=p$) or the CSA-adapted RGB features ($m=r$). Before fusion, $\mathbf{p}_p^a$ is bilinearly upsampled to the original image resolution $H \times W$. We apply a Gaussian filter $G_\sigma(\cdot)$ over both branches to mitigate noise. The final segmentation map $\mathcal{M}_{\text{final}}$ is defined as:
\begin{equation}
    \mathcal{M}_{\text{final}} = \beta \cdot G_\sigma(\mathbf{p}_r^a) + (1 - \beta) \cdot G_\sigma(\mathbf{p}_p^a)
\end{equation}
where $\beta \in [0, 1]$ is a balancing weight.

\noindent \textbf{Loss Function.}
During training on $\mathcal{D}_{\text{train}}$, we freeze the vision-language encoders and solely optimize the prompt context vectors alongside the patch adapter. We employ a composite cross-entropy classification loss $\mathcal{L}_{\text{cls}} = \mathcal{L}_{\text{cls}}^{\text{view}} + \mathcal{L}_{\text{cls}}^{\text{global}}$ to supervise both individual view-level global features and their intra-modality ensemble average. The pixel-level Visual-Language Alignment (VLA) loss $\mathcal{L}_{\text{vla}}^m$ relies on Dice~\cite{dice-loss} and Focal~\cite{focal-loss} terms to explicitly enforce spatial consistency with the binary ground-truth mask $\mathcal{M}$:
\begin{equation}
    \begin{aligned}
    \mathcal{L}_{\text{vla}}^m &= \text{Focal}([\mathbf{p}_m^n, \mathbf{p}_m^a],\; \mathcal{M}) \\
      &\quad {}+ \text{Dice}(\mathbf{p}_m^a,\; \mathcal{M}) \\
      &\quad {}+ \text{Dice}(\mathbf{p}_m^n,\; 1 - \mathcal{M})
    \end{aligned}
\end{equation}
The total objective function is formulated as:
\begin{equation}
    \mathcal{L}_{\text{total}} = \sum_{m \in \{p, r\}} \mathcal{L}_{\text{vla}}^{m} + \mathcal{L}_{\text{cls}} + \lambda \cdot \mathcal{L}_{\text{dspc}}
\end{equation}
where $\lambda$ governs the DSPC regularization scaling.

\section{Experiments}
\label{sec:experiments}

\subsection{Experimental Settings}
\label{subsec:exp_settings}

\noindent\textbf{Datasets.} We evaluate our approach on two widely used industrial anomaly detection benchmarks: the real-world MVTec 3D-AD~\cite{mvtec-3dad} and the synthetic Eyecandies~\cite{eyecandies}. Both datasets provide paired RGB images and point clouds, along with image-level labels and pixel-level masks for comprehensive evaluation.

\noindent\textbf{Evaluation Protocols.} Following established zero-shot formulations (Sec.~\ref{sec:problem_formulation}), we evaluate the generalization capability of AggCLIP under two distinct settings.
First, following the protocol established in~\cite{PointAD}, we adopt the \textbf{\textit{one-vs-rest}} setting to assess within-domain generalization: AggCLIP is trained on a single category and evaluated on the remaining unseen categories. Second, we explore a more challenging \textbf{\textit{cross-dataset}} setting to measure generalization against domain shifts. To evaluate model generalizability on unseen datasets, we train AggCLIP on MVTec 3D-AD and test it on Eyecandies.

\noindent\textbf{Evaluation Metrics.} We evaluate our method using four standard metrics. For object-level detection, we employ O-AUROC and O-AP, with the latter better reflecting performance on inherently imbalanced industrial datasets. For pixel-level localization, we utilize P-AUROC and P-AUPRO. We emphasize AUPRO as it weights distinct anomaly components equally, preventing large defects from dominating the evaluation. All metrics are reported as percentages.

\noindent\textbf{Implementation Details.} We adopt CLIP ViT-L/14@336px as the frozen multimodal backbone ($D=1024$). RGB images and rendered views are resized to $336 \times 336$. We use the Open3D library~\cite{open3d} to generate $K=9$ multi-view 2D rendered images by rotating the point cloud around the X-axis at angles of $\{-\frac{4}{5}\pi, -\frac{3}{5}\pi, \ldots, \frac{3}{5}\pi, \frac{4}{5}\pi\}$. In GSA, we set the neighborhood size strictly to $k=10$ and scaling coefficient $\gamma=0.5$. In CSA, we use multi-scale Gaussian kernel sizes $\mathcal{S} = \{1, 3, 5\}$. The standard deviation for both the CSA Gaussian kernels and the post-processing filter $G_\sigma(\cdot)$ is set to $\sigma=4$. The hidden dimension of the two-layer FFN in the Adapter is $512$. The learnable prompt context length is set to $L=12$. The temperature parameters are $\tau_{vla}=0.07$ and $\tau_{dspc}=0.5$, and the multimodal fusion weight is $\beta=0.5$. Training runs for 15 epochs with Adam (batch size 4, learning rate $0.001$, $\beta_1=0.5$, $\beta_2=0.999$), and the DSPC loss weight is $\lambda=1$. All experiments are implemented in PyTorch on a single RTX 3090 GPU. More details are available in the Appendix.

\begin{table}[t]
\centering
\caption{Performance comparison of zero-shot multimodal anomaly detection under the \textit{one-vs-rest} setting on the MVTec 3D-AD and Eyecandies datasets. Best results are highlighted in \textcolor{red}{red}, and runners-up in \textcolor{blue}{blue}. Throughout all tables, O-R, O-A, P-R, and P-P denote O-AUROC, O-AP, P-AUROC, and P-AUPRO, respectively.}
\label{tab:zs_m3d_cross_category}
\small
\setlength{\tabcolsep}{2.2pt}
\renewcommand{\arraystretch}{1.1} 
\resizebox{\columnwidth}{!}{
\begin{tabular}{lcccc}
\toprule
& \multicolumn{2}{c}{\textbf{MVTec 3D-AD}} & \multicolumn{2}{c}{\textbf{Eyecandies}} \\
\cmidrule(lr){2-3} \cmidrule(lr){4-5}
\textbf{Methods} & \textbf{Object-level} & \textbf{Pixel-level} & \textbf{Object-level} & \textbf{Pixel-level} \\
& \textbf{(O-R, O-A)} & \textbf{(P-R, P-P)} & \textbf{(O-R, O-A)} & \textbf{(P-R, P-P)} \\
\midrule
CLIP~\cite{CLIP}           & (60.4, 86.4) & (--  , 56.0) & (73.0, 73.9) & (78.0, 31.8) \\
AnomalyCLIP~\cite{AnomalyCLIP}    & (66.2, 87.6) & (91.6, 70.9) & (65.0, 67.5) & (85.0, 56.2) \\
PointAD~\cite{PointAD}            & (86.9, 96.1) & (97.2, 90.2) & (77.7, 80.4) & (95.3, 84.3) \\
ZUMA-FT~\cite{ZUMA}               & (\textcolor{blue}{88.2}, --  ) & (\textcolor{blue}{97.8}, \textcolor{blue}{93.1}) & (78.2, --  ) & (\textcolor{blue}{97.5}, \textcolor{blue}{89.4}) \\
GS-CLIP~\cite{GS-CLIP}            & (\textcolor{blue}{88.2}, \textcolor{red}{97.5}) & (97.6, 91.3) & (\textcolor{blue}{79.3}, \textcolor{blue}{82.2}) & (95.8, 86.4) \\
\rowcolor{gray!15}
\textbf{AggCLIP (Ours)} & (\textcolor{red}{88.3}, \textcolor{red}{97.5}) & (\textcolor{red}{98.4}, \textcolor{red}{93.9}) & (\textcolor{red}{79.8}, \textcolor{red}{82.4}) & (\textcolor{red}{98.2}, \textcolor{red}{89.8}) \\
\bottomrule
\end{tabular}
}
\end{table}

\begin{table}[t]
\centering
\caption{Performance comparison of zero-shot multimodal anomaly detection under the \textit{cross-dataset} setting (trained on MVTec 3D-AD, tested on Eyecandies). * Indicates results obtained by running the official code.}
\label{tab:zs_m3d_cross_dataset}
\small 
\setlength{\tabcolsep}{8pt}
\renewcommand{\arraystretch}{1.1}
\begin{tabular}{l cc}
\toprule
& \textbf{Object-level} & \textbf{Pixel-level} \\
\multirow{-2}{*}{\textbf{Methods}} & \textbf{(O-R, O-A)} & \textbf{(P-R, P-P)} \\
\midrule
AnomalyCLIP~\cite{AnomalyCLIP} & (65.7, 68.1) & (86.2, 61.3) \\
PointAD~\cite{PointAD}         & (78.6, 80.8) & (94.0, 80.7) \\
ZUMA-FT*~\cite{ZUMA}           & (79.1, 81.2)   & (\textcolor{blue}{94.8}, 81.9) \\
GS-CLIP*~\cite{GS-CLIP}        & (\textcolor{blue}{80.2}, \textcolor{blue}{82.2}) & (94.6, \textcolor{blue}{82.4}) \\
\rowcolor{gray!15}
\textbf{AggCLIP (Ours)}       & (\textcolor{red}{81.4}, \textcolor{red}{84.2}) & (\textcolor{red}{95.8}, \textcolor{red}{85.3})\\
\bottomrule
\end{tabular}
\end{table}

\subsection{Main Experimental Results}

\noindent\textbf{Quantitative Analysis.} Tables~\ref{tab:zs_m3d_cross_category} and~\ref{tab:zs_m3d_cross_dataset} demonstrate that AggCLIP achieves state-of-the-art zero-shot performance under both the \textit{one-vs-rest} and \textit{cross-dataset} settings. In the \textit{one-vs-rest} evaluation, AggCLIP consistently ranks first across all reported metrics for both the MVTec 3D-AD and Eyecandies. Specifically, on MVTec 3D-AD, it outpaces PointAD~\cite{PointAD} by 1.4\% in O-AUROC and 3.7\% in P-AUPRO, while also achieving more precise pixel-level localization than strong baselines like GS-CLIP~\cite{GS-CLIP} and ZUMA-FT~\cite{ZUMA}. On the Eyecandies dataset, AggCLIP establishes new peaks of 79.8\%/82.4\% in O-AUROC/O-AP at the object level and 98.2\%/89.8\% in P-AUROC/P-AUPRO at the pixel level. Furthermore, under the more challenging \textit{cross-dataset} setting, AggCLIP maintains its superiority against all competitors. It surpasses the runner-up by absolute margins of 1.2\% and 2.0\% in O-AUROC and O-AP over GS-CLIP~\cite{GS-CLIP}, alongside notable improvements of 1.0\% in P-AUROC over ZUMA-FT~\cite{ZUMA} and 2.9\% in P-AUPRO over GS-CLIP, successfully verifying its stronger generalization against domain shifts. More quantitative results are available in the Appendix.

\begin{figure}[t]
    \centering
    \includegraphics[width=\linewidth]{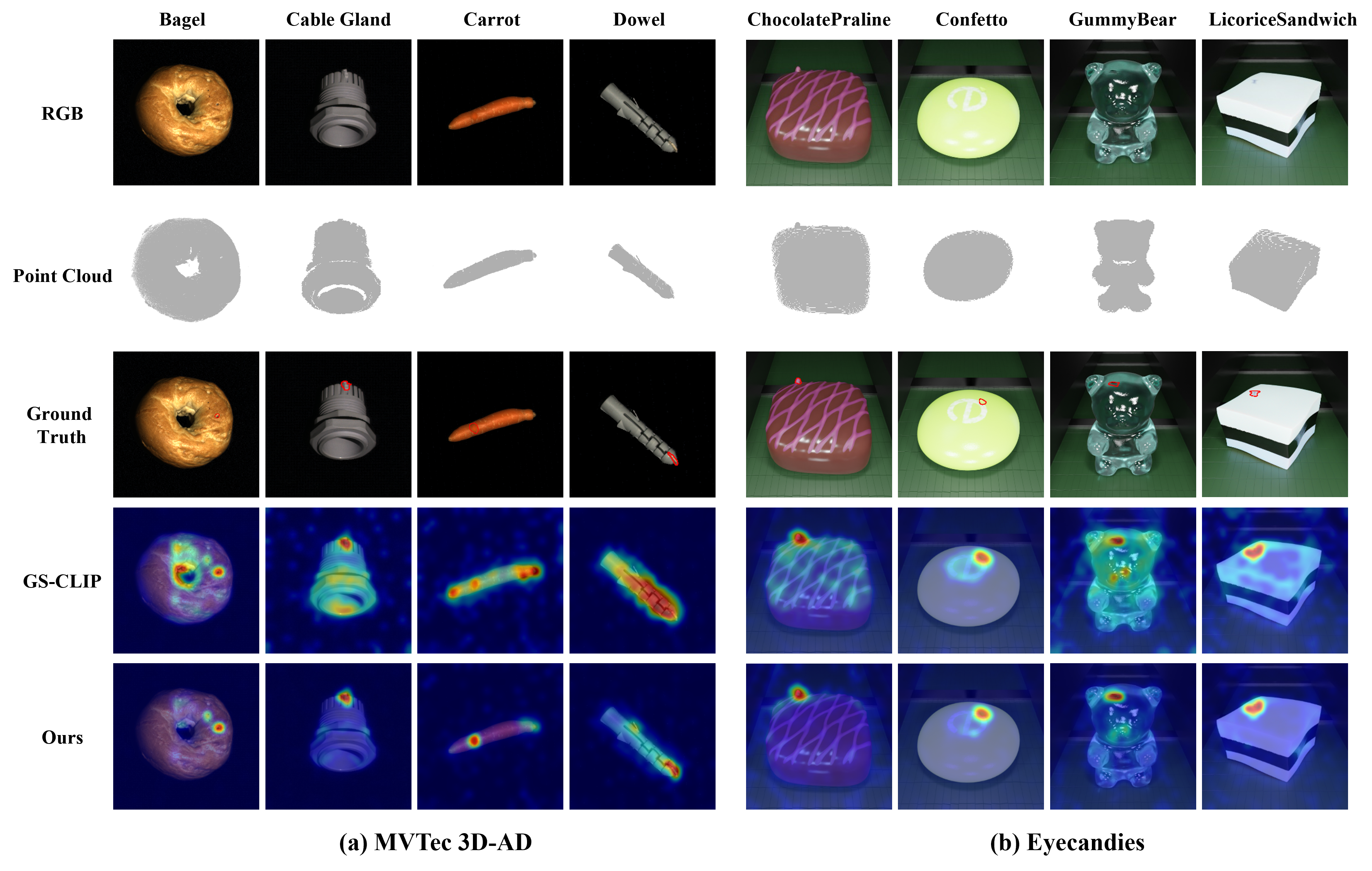}
    \caption{Qualitative results on MVTec 3D-AD and Eyecandies. From top to bottom: input RGB images, 3D point clouds, ground truth, anomaly maps from GS-CLIP, and anomaly maps from AggCLIP.}
    \Description{Qualitative examples from MVTec 3D-AD and Eyecandies are arranged by sample. Rows show the RGB input, point cloud, ground-truth defect mask, GS-CLIP anomaly map, and AggCLIP anomaly map. AggCLIP more closely follows the defect boundaries and suppresses background responses.}
    \label{fig:qualitative}
\end{figure}

\noindent\textbf{Qualitative Analysis.} 
Figure~\ref{fig:qualitative} shows that AggCLIP produces anomaly maps with clearer boundaries and fewer false activations than GS-CLIP~\cite{GS-CLIP}. It localizes subtle or irregular defective regions more completely, while suppressing responses on normal textured areas. We attribute this improvement to our spatial aggregation modules (GSA and CSA), which incorporate local geometric structure and appearance spatial context into the frozen CLIP features, enabling more fine-grained and spatially coherent anomaly localization. This indicates that aggregating appearance and geometric information helps the model better distinguish true defects from category-specific visual variations. More qualitative results are available in the Appendix.

\begin{table}[t]
\centering
\caption{Efficiency comparison of different methods. * Indicates results obtained by running the official code.}
\label{tab:efficiency}
\small
\setlength{\tabcolsep}{5pt}
\renewcommand{\arraystretch}{1.2}
\resizebox{\columnwidth}{!}{
\begin{tabular}{l cccc}
\toprule
\textbf{Methods} & \textbf{GFLOPs} & \textbf{FPS} & \textbf{Time (s)} & \textbf{Memory (MB)} \\
\midrule
CLIP*~\cite{CLIP}     & 2330.25 & 3.64 & 0.27 & 2270 \\
PointAD*~\cite{PointAD}          & 4252.45          & 3.20          & 0.31          & 2610         \\
ZUMA-FT*~\cite{ZUMA}             & 4469.10          & 0.53          & 1.89          & 3856         \\
GS-CLIP*~\cite{GS-CLIP}       & 9591.98          & 1.12          & 0.89          & 3677\\
\rowcolor{gray!20}
\textbf{AggCLIP (Ours)}                   & 4259.97 & 3.07 & 0.33 & 2645 \\
\bottomrule
\end{tabular}
}
\end{table}

\subsection{Computational Efficiency}

Because inference speed and resource management are critical for real-world industrial deployment, we evaluated the computational overhead of various methods, as detailed in Table~\ref{tab:efficiency}. AggCLIP consistently outperforms both ZUMA-FT~\cite{ZUMA} and GS-CLIP~\cite{GS-CLIP} across all efficiency metrics. Although it is marginally slower than PointAD~\cite{PointAD}, AggCLIP yields substantial gains in anomaly detection and localization accuracy. This demonstrates that AggCLIP achieves an optimal trade-off between inference speed and detection performance.

\subsection{Ablation Studies}

\begin{table}[t]
\centering
\caption{Ablation on different components.}
\label{tab:ablation-components}
\small
\renewcommand{\arraystretch}{1.1}
\resizebox{\columnwidth}{!}{
\begin{tabular}{ccccc cc}
\toprule
& & & & & \textbf{Object-level} & \textbf{Pixel-level} \\
\multirow{-2}{*}{\textbf{Version}} & \multirow{-2}{*}{\textbf{GSA}} & \multirow{-2}{*}{\textbf{CSA}} & \multirow{-2}{*}{\textbf{Adapter}} & \multirow{-2}{*}{\textbf{$\mathcal{L}_{\text{dspc}}$}} & \textbf{(O-R, O-A)} & \textbf{(P-R, P-P)} \\
\midrule
V1 & \ding{55} & \ding{55} & \ding{55} & \ding{55} & (81.2, 89.6) & (90.3, 88.9)   \\
V2 & \ding{51} & \ding{55} & \ding{55} & \ding{55} & (87.4, 96.7) & (97.8, 93.0) \\
V3 & \ding{55} & \ding{51} & \ding{55} & \ding{55} & (85.5, 94.3) & (95.2, 90.4) \\
V4 & \ding{55} & \ding{51} & \ding{51} & \ding{55} & (86.8, 95.7) & (96.2, 91.3) \\
V5 & \ding{51} & \ding{51} & \ding{51} & \ding{55} & (\textcolor{blue}{87.7}, \textcolor{blue}{97.1}) & (\textcolor{blue}{98.0}, \textcolor{blue}{93.6}) \\
\rowcolor{gray!15}
V6 & \ding{51} & \ding{51} & \ding{51} & \ding{51} & (\textcolor{red}{88.3}, \textcolor{red}{97.5}) & (\textcolor{red}{98.4}, \textcolor{red}{93.9}) \\
\bottomrule
\end{tabular}%
}
\end{table}

\textbf{Ablation on Different Components.} 
Table~\ref{tab:ablation-components} evaluates the contribution of each proposed component. The baseline (V1) retains only the dual-branch architecture with decoupled learnable prompts, omitting GSA, CSA, the Adapter, and $\mathcal{L}_{\text{dspc}}$. Notably, integrating GSA into the 3D branch (V2) yields the most substantial single-module improvement (+6.2\% O-AUROC, +4.1\% P-AUPRO), confirming that density-adaptive geometric aggregation is critical for stabilizing noisy back-projected point features. A finer-grained comparison of different aggregation strategies within GSA is provided in the Appendix. Conversely, in the RGB branch, applying CSA (V3) and further equipping the Adapter (V4) progressively boost performance. The Adapter contributes an additional +1.3\% O-AUROC and +0.9\% P-AUPRO over CSA alone, validating the necessity of learnable feature adaptation for 2D appearance features. When the spatial aggregation modules from both modalities are jointly combined (V5), strong 2D--3D complementarity emerges, pushing P-AUROC to 98.0\%. Finally, the full framework (V6), augmented with the $\mathcal{L}_{\text{dspc}}$ loss, achieves the highest performance across all metrics, verifying that explicit cross-modal prompt alignment effectively tightens semantic consistency.

\begin{figure}[t]
\centering
\begin{subfigure}[b]{\columnwidth}
  \centering
  \includegraphics[width=\columnwidth]{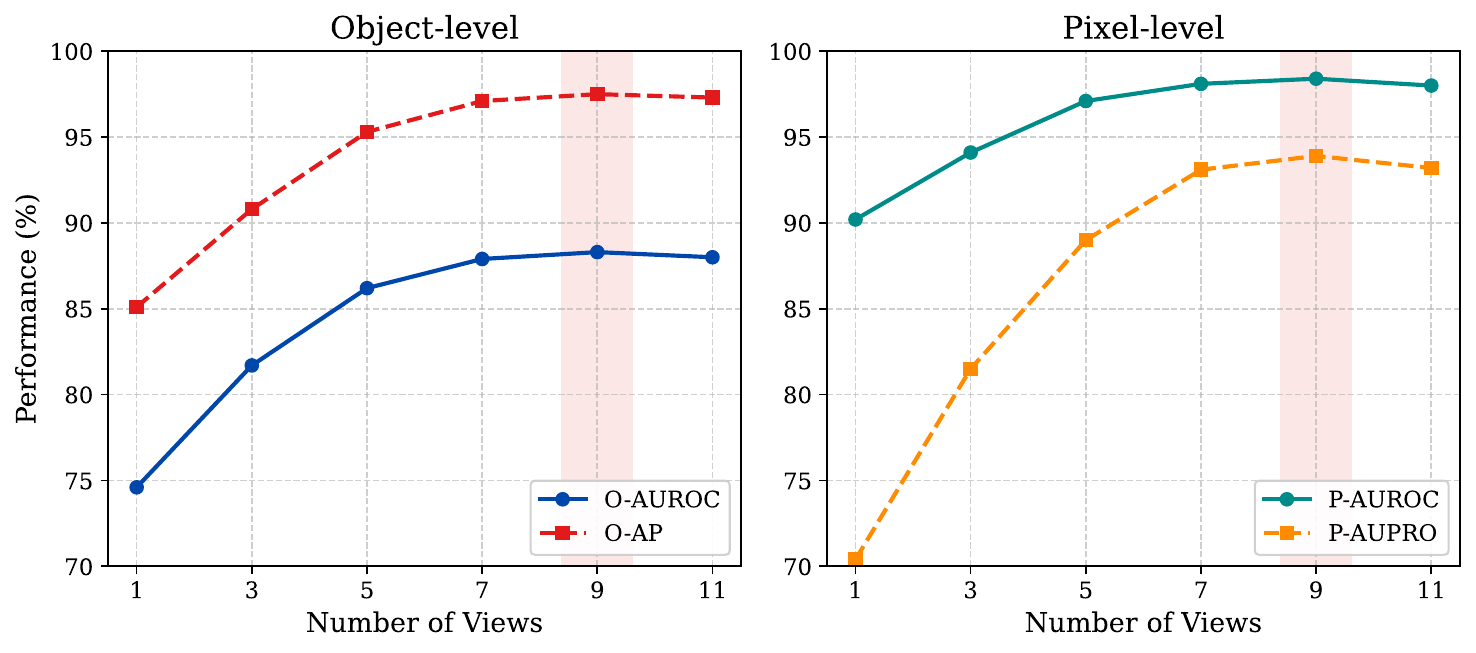}
  \caption{Ablation on the rendering view number $K$.}
  \label{fig:ablation_view_number}
\end{subfigure}
\vspace{6pt}
\begin{subfigure}[b]{\columnwidth}
  \centering
  \includegraphics[width=\columnwidth]{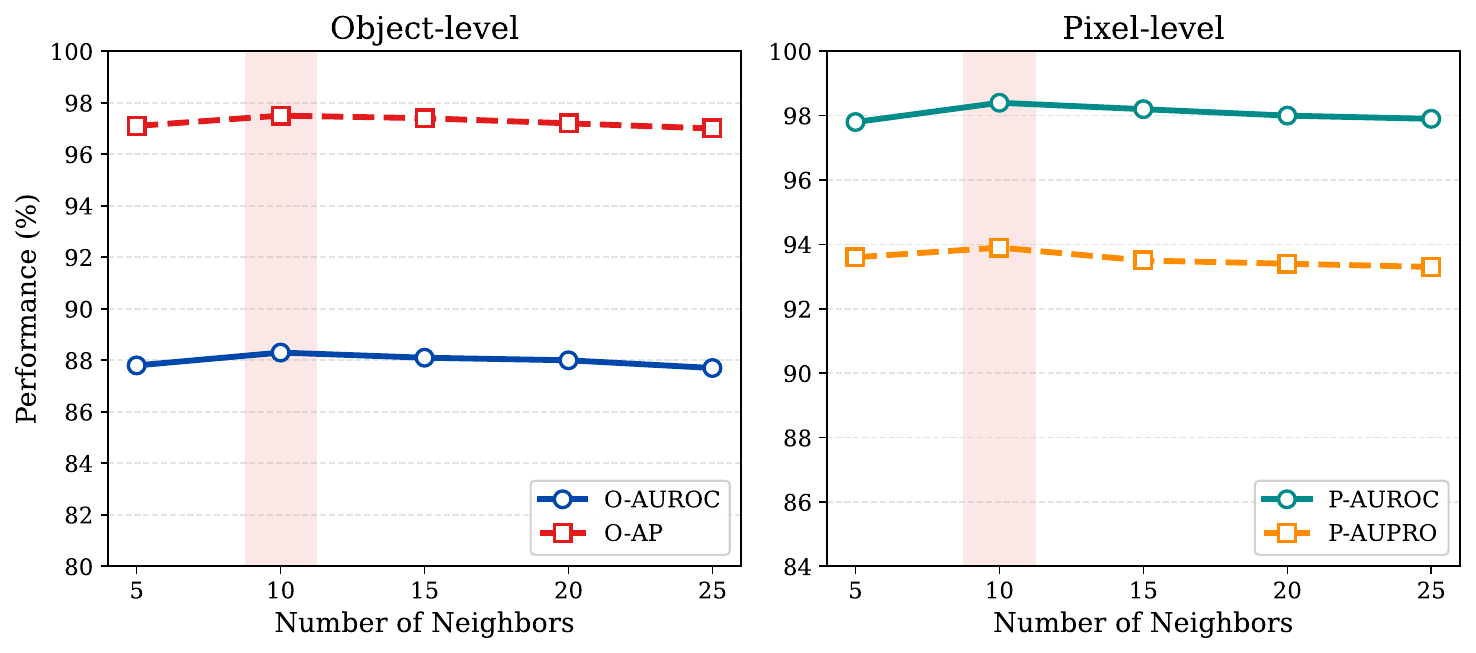}
  \caption{Ablation on the number of neighbors $k$.}
  \label{fig:ablation_neighbor_number}
\end{subfigure}
\caption{\textbf{Ablation on key hyperparameters on MVTec 3D-AD under the \textit{one-vs-rest} setting.} (a) Varying the number of rendered views $K$. (b) Varying the GSA neighborhood size $k$.}
\Description{Two line charts report object- and pixel-level anomaly detection metrics. Performance increases with the number of rendered views and saturates at nine views, while the neighborhood-size study reaches its best overall result at ten neighbors and remains stable around that setting.}
\label{fig:ablation-views-neighbors}
\end{figure}

\noindent\textbf{Ablation on the Number of Rendering Views.} The number of rendered views $K$ controls the richness of 3D structural information transferred to point features. Figure~\ref{fig:ablation-views-neighbors}(a) demonstrates that increasing $K$ steadily improves performance by providing more comprehensive multi-view geometric information. Performance saturates at $K=9$, which we adopt to balance accuracy and efficiency.

\noindent\textbf{Ablation on the Number of Neighbors in GSA.} Figure~\ref{fig:ablation-views-neighbors}(b) reports GSA performance as a function of the neighborhood size $k$. Both object- and pixel-level metrics peak at $k=10$, which we adopt as the default. Performance varies only marginally across $k \in \{5, 10, 15, 20, 25\}$, confirming that GSA is stable under the choice of neighborhood scale.

\begin{table}[t]
\centering
\caption{Ablation study on the multi-scale Gaussian kernel sizes $\mathcal{S}$ in the CSA module.}
\label{tab:ablation-csa-kernel}
\small 
\setlength{\tabcolsep}{8pt}
\renewcommand{\arraystretch}{1.1}
\begin{tabular}{lcc}
\toprule
& \textbf{Object-level} & \textbf{Pixel-level} \\
\multirow{-2}{*}{\textbf{Kernel Sizes}} & \textbf{(O-R, O-A)} & \textbf{(P-R, P-P)} \\
\midrule
$\{1\}$      & (88.0, 97.1) & (97.3, 92.9) \\
$\{3\}$      & (87.9, 96.8) & (97.7, 92.6) \\
$\{5\}$      & (87.7, 97.2) & (97.8, 93.0) \\
$\{1, 3\}$   & (\textcolor{blue}{88.1}, \textcolor{blue}{97.4}) & (\textcolor{blue}{98.1}, \textcolor{blue}{93.5}) \\
\rowcolor{gray!20}
$\{1, 3, 5\}$ & (\textcolor{red}{88.3}, \textcolor{red}{97.5}) & (\textcolor{red}{98.4}, \textcolor{red}{93.9}) \\
\bottomrule
\end{tabular}
\end{table}

\noindent\textbf{Ablation on the Multi-Scale Gaussian Kernels in CSA.} 
As shown in Table~\ref{tab:ablation-csa-kernel}, we investigate the influence of the Gaussian kernel set $\mathcal{S}$ in the CSA module. While single-scale kernels provide a baseline, they suffer from a limited receptive field. Specifically, $\{1\}$ captures fine details but lacks spatial smoothing, whereas $\{3\}$ and $\{5\}$ aggregate broader context at the expense of subtle defect boundaries. Combining these scales establishes a strong hierarchical representation. Integrating $\{1, 3\}$ effectively bridges local and intermediate features, yielding notable improvements. Ultimately, the full multi-scale setting $\{1, 3, 5\}$ achieves optimal performance across all metrics, peaking at 88.3\% O-AUROC and 98.4\% P-AUROC. This validates that combining distinct kernel sizes allows the CSA module to simultaneously preserve fine-grained structural integrity and encode rich regional context for accurate anomaly localization.

\noindent\textbf{Ablation on Modality.} 
To evaluate the necessity of multimodal inputs, Table~\ref{tab:ablation-modality} compares single-modality baselines against our full framework. Relying solely on either 2D RGB or 3D point cloud data yields suboptimal performance. In contrast, by jointly fusing 2D and 3D features, AggCLIP achieves substantial gains across both the MVTec 3D-AD and Eyecandies datasets. This confirms that integrating complementary multimodal information is crucial for effective anomaly detection. Ablations on hyperparameter sensitivity and rendering condition are available in the Appendix.

\begin{table}[t]
\centering
\caption{\textbf{Ablation on modality.}}
\label{tab:ablation-modality}
\small
\setlength{\tabcolsep}{2pt}
\renewcommand{\arraystretch}{1.1}
\begin{tabular}{@{}lccc@{}}
\toprule
& & \textbf{Object-level} & \textbf{Pixel-level} \\
\multirow{-2}{*}{\textbf{Dataset}} & \multirow{-2}{*}{\textbf{Modality}} & \textbf{(O-R, O-A)} & \textbf{(P-R, P-P)} \\
\midrule
\multirow{3}{*}{MVTec 3D-AD}
& RGB                 & (82.3, 88.2) & (93.4, 86.7) \\
& Point Cloud                 & (\textcolor{blue}{84.1}, \textcolor{blue}{96.6}) & (\textcolor{blue}{97.2}, \textcolor{blue}{88.3}) \\
\rowcolor{gray!15}
\cellcolor{white} & \shortstack{RGB +\\Point Cloud} & (\textcolor{red}{88.3}, \textcolor{red}{97.5}) & (\textcolor{red}{98.4}, \textcolor{red}{93.9}) \\
\midrule
\multirow{3}{*}{Eyecandies}
& RGB                 & (\textcolor{blue}{74.5}, 77.3) & (\textcolor{blue}{93.6}, 78.8) \\
& Point Cloud                 & (73.7, \textcolor{blue}{78.5}) & (92.4, \textcolor{blue}{79.6}) \\
\rowcolor{gray!15}
\cellcolor{white} & \shortstack{RGB +\\Point Cloud} & (\textcolor{red}{79.8}, \textcolor{red}{82.4}) & (\textcolor{red}{98.2}, \textcolor{red}{89.8}) \\
\bottomrule
\end{tabular}
\end{table}

\section{Conclusion}
\label{sec:conclusion}
In this paper, we presented \textbf{AggCLIP}, a novel zero-shot multimodal anomaly detection framework that adapts a frozen CLIP backbone for dense industrial inspection. To bridge the gap between image-level pre-training and fine-grained defect localization, AggCLIP explicitly synergizes 2D appearance and 3D geometric information through modality-specific spatial aggregation and aligns them via DSPC learning. This design effectively overcomes the global bias inherent in pre-trained vision-language models. Extensive evaluations demonstrate that AggCLIP establishes new state-of-the-art performance on the MVTec 3D-AD and Eyecandies datasets across both one-vs-rest and cross-dataset settings, all while maintaining high inference efficiency.

\section*{Acknowledgments}
This work was supported by the Beijing Natural Science Foundation under Grant No. L243018.

\input{AggCLIP_OneThree.bbl}

\end{document}